\documentclass[10pt,twocolumn,letterpaper]{article}

\usepackage{cvpr}      

\usepackage{orcidlink}
\def\skrylle{Spruce~A}
\def\snoge{Spruce~B}
\def\HH{\operatorname{H}}
\def\FF{\operatorname{F }}
\usepackage{amsfonts}
\usepackage{amsmath}
\usepackage{amssymb}
\usepackage{amsthm}
\usepackage{bbm}
\usepackage{overpic} 
\usepackage{float}

\definecolor{cvprblue}{rgb}{0.21,0.49,0.74}

\def\paperID{} 
\def\confName{}
\def\confYear{}

\title{The Impact of Semi-Supervised Learning on Line Segment Detection}

\author{Johanna Engman \qquad Karl Åström \qquad Magnus Oskarsson\\
CVML, Centre for mathematical sciences, Lund University, Sweden\\
{\tt\small \{johanna.engman, karl.astrom, magnus.oskarsson\}@math.lth.se}
}

\begin{document}
\maketitle
\begin{abstract}
In this paper we present a method for line segment detection in images, based on a semi-supervised framework. Leveraging the use of a  consistency loss based on differently augmented and perturbed  unlabeled images with a small amount of labeled data, we show comparable results to fully supervised methods. This opens up application scenarios where annotation is difficult or expensive, and for domain specific adaptation of models. We are specifically interested in real-time and online applications, and investigate small and efficient learning backbones. Our method is to our knowledge the first to target line detection using modern state-of-the-art methodologies for semi-supervised learning. We test the method on both standard benchmarks and domain specific scenarios for forestry applications, showing the tractability of the proposed method. Code: \texttt{https://github.com/jo6815en/semi-lines/}
\end{abstract}   
\section{Introduction}
\label{sec:intro}

\begin{figure*}
 \begin{tabular}{cccc}
 ~~~~~~~~~~~Ground Truth~~~~ & ~~~~~~~~~~~~~~~Fully Supervised~~ & ~~~~~~~~~~~~~~~Fully Supervised~~ & ~~~~~~~~~~~~~~~~~~~Proposed~~~~~~ \\
& ~~~~~~~~~~~~(1/16 data) & ~~~~~~~~~~~~(All data) & ~~~~~~~~~~~~(1/16 labeled data)
 \end{tabular}\\
  \includegraphics[width=.24\textwidth]{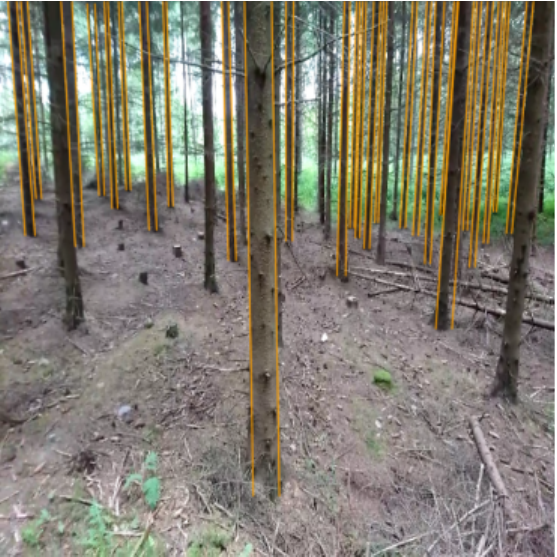}
  \includegraphics[width=.24\textwidth]{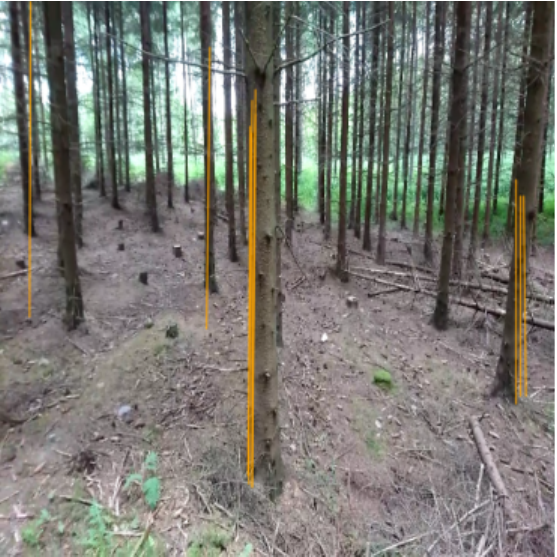}
    \includegraphics[width=.24\textwidth]{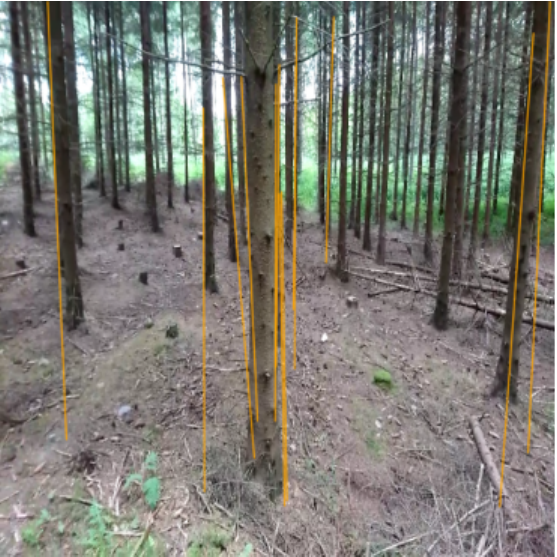}
   \includegraphics[width=.24\textwidth]{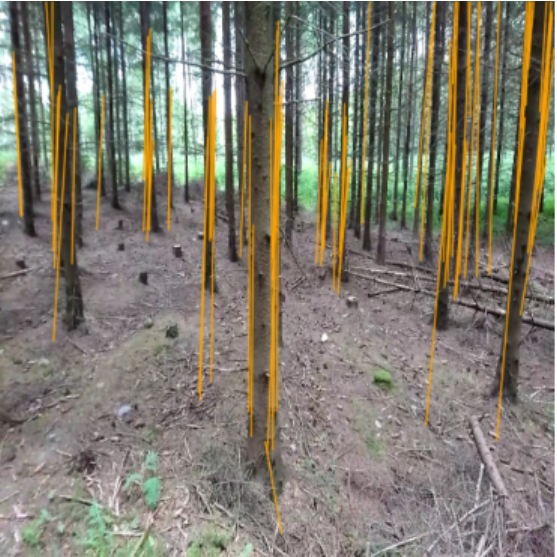}
    \caption{From left to right: Ground truth detection, supervised model on 1/16 of data, supervised model on all data, proposed semi-supervised approach with 1/16 labeled data. The semi-supervised method improves significantly on the supervised method. }
    \label{fig:comparison1}
\end{figure*}

Line segments are prevalent in images depicting scenes of both in- and outdoor human-made environments as well as in nature. Line structures are used by humans for efficient and robust interpretation of the world  \cite{cole2023people,biederman1988surface}, and can also play a crucial role in automatic  scene comprehension, encapsulating scene structure efficiently \cite{zhou2019learning}. Applications and down stream tasks that are made more tractable encompass various computer vision tasks, including 3D reconstruction, Structure-from-Motion (SfM) \cite{hofer2017efficient,mateus2021incremental,micusik2017structure}, Simultaneous Localization and Mapping (SLAM) \cite{gomez2019pl}, visual localization \cite{gao2022pose}, tracking, and vanishing point estimation \cite{tardif2009non}. Lines can also be efficient intermediate representations for subsequent semantic processing. 
Compared to point based features, line features are more stable over time (this is true both on a small and large time scale) and more invariant to changes in scene environment such as \eg lighting, season and weather conditions as shown in \cite{sattler2018benchmarking}. 

Although lines are perhaps mostly associated with man-made structures, there are many examples of linear structures in nature. In this paper we use, as a domain specific scenario, forestry images. Detecting lines in tree imagery is challenging, and many applications related to them are of particular importance. From an environmental stand-point trees can contribute with both carbon-dioxide binding and new sustainable materials in diverse industries. For many automation use cases involving forestry we would like to monitor, measure and interpret forestry scenes, which translates to both semantic computer vision tasks such as classification and scene understanding and geometric computer vision tasks such as camera pose estimation and tracking. 
\begin{figure*}
\begin{tabular}{ccc}
~~~~~~~~~~~~~~~~~~~~~~DeepLSD~~~~~~~~~~~~ & ~~~~~~~~~~~~~~~~~~~~~~~~~~~~~~~~~~~~~LETR~~~~~~~~~~~~~ & ~~~~~~~~~~~~~~~~~~~~~~~~~~~~~~~~~~~~~~M-LSD~~~~~~~~~~~~~
\end{tabular}\\
  \includegraphics[width=1.0\textwidth]{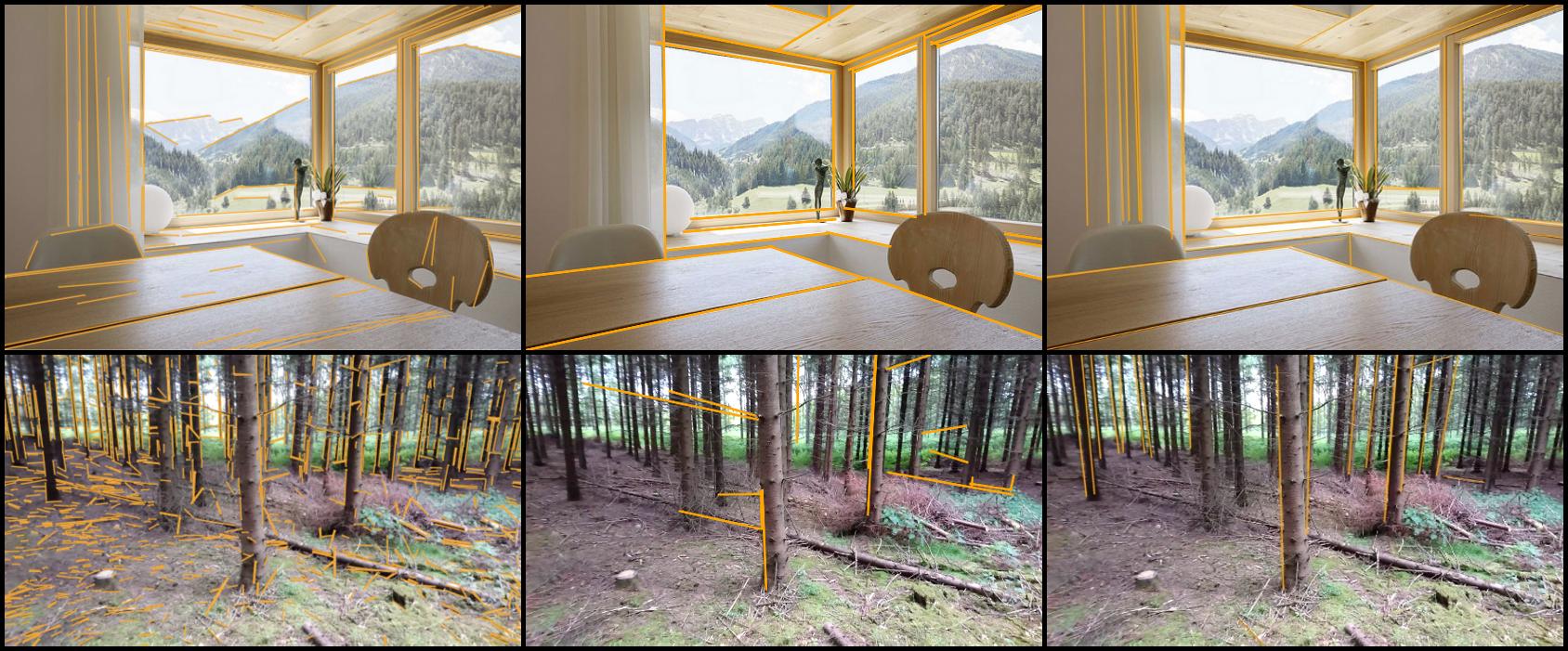}
    \caption{Many trained models fail to generalize to new image domains. Top row shows the output from left to right:  DeepLSD \cite{pautrat2023deeplsd}, LETR \cite{xu2021line}, M-LSD \cite{mlsd}, trained on the Wireframe dataset. Bottom row shows the output from these models on an out of domain image.}
    \label{fig:badgeneral}
    \end{figure*}

Many newly formulated learned line segment detectors have shown both promising and impressive results \cite{pautrat2023deeplsd,xu2021line,mlsd}. However, many such large models fail to transition and generalize to new image domains without retraining, see Fig.~\ref{fig:badgeneral}.



To tackle the expensive need for annotations and in contrast to large general and slower models, we propose a method for semi-supervised line segment detection. Our method allows for domain specific and light-weight processing. At the same time, our use of semi-supervision enables better generalization and generates a method that better adapts to new scenes. We build on the ideas of semi-supervised learning for semantic segmentation from \cite{fixmatch} and \cite{unimatch}. Combining these ideas with a compact line segment model \cite{mlsd} we propose the first framework for semi-supervised line segment detection. This results in a fast, efficient, and accurate system that is trained on a smaller portion of the whole dataset and performs as well or even better than the standard supervised method.

As there are no other semi-supervised frameworks for line segment detection, we perform comparison between our models and the supervised version of our model. Exploring the fundamental possibility of gaining information and boosting the performance with a lower amount of annotated data. We do however include the performance of pre-trained DeepLSD \cite{pautrat2023deeplsd} and LETR \cite{xu2021line} models for reference.

We show that our method  learns consistency  information from unlabeled images together with task specific information from annotated images. For our forestry use case, only having annotations of half the dataset but using our method results in 28\% higher score in sAP$^{10}$ compared to using a fully supervised method with all the data annotated.







Our main contributions are:
\begin{enumerate}
    \item We present the first framework for semi-supervised learning for line segment detection. 
\item We show that semi-supervised methods work well for line segment detection, improving significantly on the state-of-the-art for new environments, where there is little annotation. 
\item We show that small models, suitable for real-time applications, work well and can heavily benefit from semi-supervised methods.
\end{enumerate}
Additionally we provide two new datasets for line segment detection containing forest scenes with ground truth lines. 
We will publicly release and publish all code and data. 

\section{Related work}
In this paper we develop the first semi-supervised framework for line-segment detection. Here we provide background on line detection in general (Section~\ref{sec:related_lines}) and on semi-supervised learning (Section~\ref{sec:related_semisupervised}). 

\subsection{Line detection} \label{sec:related_lines}
Early methods for detecting line-segments rely on detected edge features based either on first order methods, e.g.\ finding local maxima to smoothed image gradients \cite{marr1980theory,canny1986computational} or second order methods, e.g.\ finding zero crossings of the laplacian, \cite{haralick1982zero}. Edge features are then typically grouped into line segments, e.g.\ using the Hough transform, \cite{hough1962method,elder2020mcmlsd} or line segment growing, \cite{von2008lsd,akinlar2011edlines}.

Over the last years, as in most fields, learning based models have increased the performance of line segment detection.
Some recent well functioning detectors are the transformer based  LETR \cite{xu2021line} and DeepLSD \cite{pautrat2023deeplsd} that incorporates an optimization based refinement module for higher accuracy. 
A similar task to line segment detection is creating systems 
that directly identify wireframes, both lines and junctions. 
In End-to-End Wireframe Parsing \cite{end} the authors focus on this problem.  In addition to their model, they introduced a novel evaluation metric, sAP - structural average precision. This metric is more robust against overlapping line segments and multiple line detections for the same line. sAP is defined as the area under the precision recall curve computed from a scored list of the detected line segments. 
To evaluate our  line segment detection we also use this metric in this paper.
This  method and other end-to-end wireframe methods, e.g.\ \cite{huang2018learning,gillsjo2023iccv,lin2020deep,xue2019learning,xue2020holistically} have spurred new interest in more general line detection algorithms, \cite{gu2021towards}.

We build our system using the line segment detector M-LSD proposed in \cite{mlsd}, since it is fast and light-weight. Their main focus was creating a framework that is possible to run in real-time on a mobile device, removing multi-module processes to simplify the model. A byproduct of these achievements is the smaller need for data, which fits very well with our goals. To optimize the performance level, \cite{mlsd} present segment-of-line (SoL) augmentations combined with a matching and a geometric loss. Their main contribution is their SoL-augmentations, where they divide a line into overlapping segments that should all align. They use the Tri-point (TP) representation \cite{tp}. Another change from earlier work is that M-LSD generates line segments directly from the final feature maps in a single module process.



\subsection{Semi-supervised learning} \label{sec:related_semisupervised}

Semi-supervised learning is a field of machine learning that tries to overcome the large need for annotated data and labels, combining the information gained in a smaller portion of labeled data together with the possibility to train on a larger dataset without labels. One of the first frameworks that succeeded with this was Mean Teacher \cite{meanteacher}, where they use two models simultaneously, one student learning fast both with labels and consistency towards the teacher,  that learns slower over an average of the student. Mean Teacher uses the fact that augmenting the inputs differently should still generate the same label.

In 2018 \cite{semiseg_2018} proposed an adversarial learning approach for semi-supervised learning for semantic segmentation, building on the ideas of GANS. Since then a number of improvements have been made, mostly abandoning the GAN-influenced ideas for contrastive learning  and Mean Teacher approaches \cite{chen2020simple,alonso2021semi,1640964}. 

More closely related to our approach is the
FixMatch model, which is a semi-supervised framework for classification using the weak-to-strong consistency regularization \cite{fixmatch}. It uses one strongly augmented image together with one weakly augmented image to obtain information without labels. They propose to use a shared model, unlike the Mean Teacher method. This model has been further developed, creating more advanced frameworks \cite{u2pl}.

Recently, a new semi-supervised learning framework, UniMatch,  for semantic segmentation  was proposed \cite{unimatch}. In contrast to recent development, they go back to the, fundamentally, simpler method of FixMatch \cite{fixmatch}, but add a second strongly perturbed version of the unlabeled input image.
They use two strong views that simultaneously are guided by a common weak view. The role of the strong perturbed images is to minimize the distance between them, similar to contrastive learning. In \cite{unimatch} they perform extensive ablation studies showing that two strongly perturbed views achieve higher  accuracy, compared to using one or three. 
 In addition they add a version of the weakly augmented image and add an augmentation in the feature domain.
For the unlabeled images \cite{unimatch} uses CutMix \cite{cutmix} as a regularization technique. While this is a good idea for classification and segmentation, we found that it is not an ideal technique for line detection. Using CutMix counteracts the system from finding long consistent lines, and we propose a variant more suited to line detection.

\begin{figure*}
\centering
 \resizebox{0.8\textwidth}{!}{
\begin{overpic}[width=1.4\columnwidth]{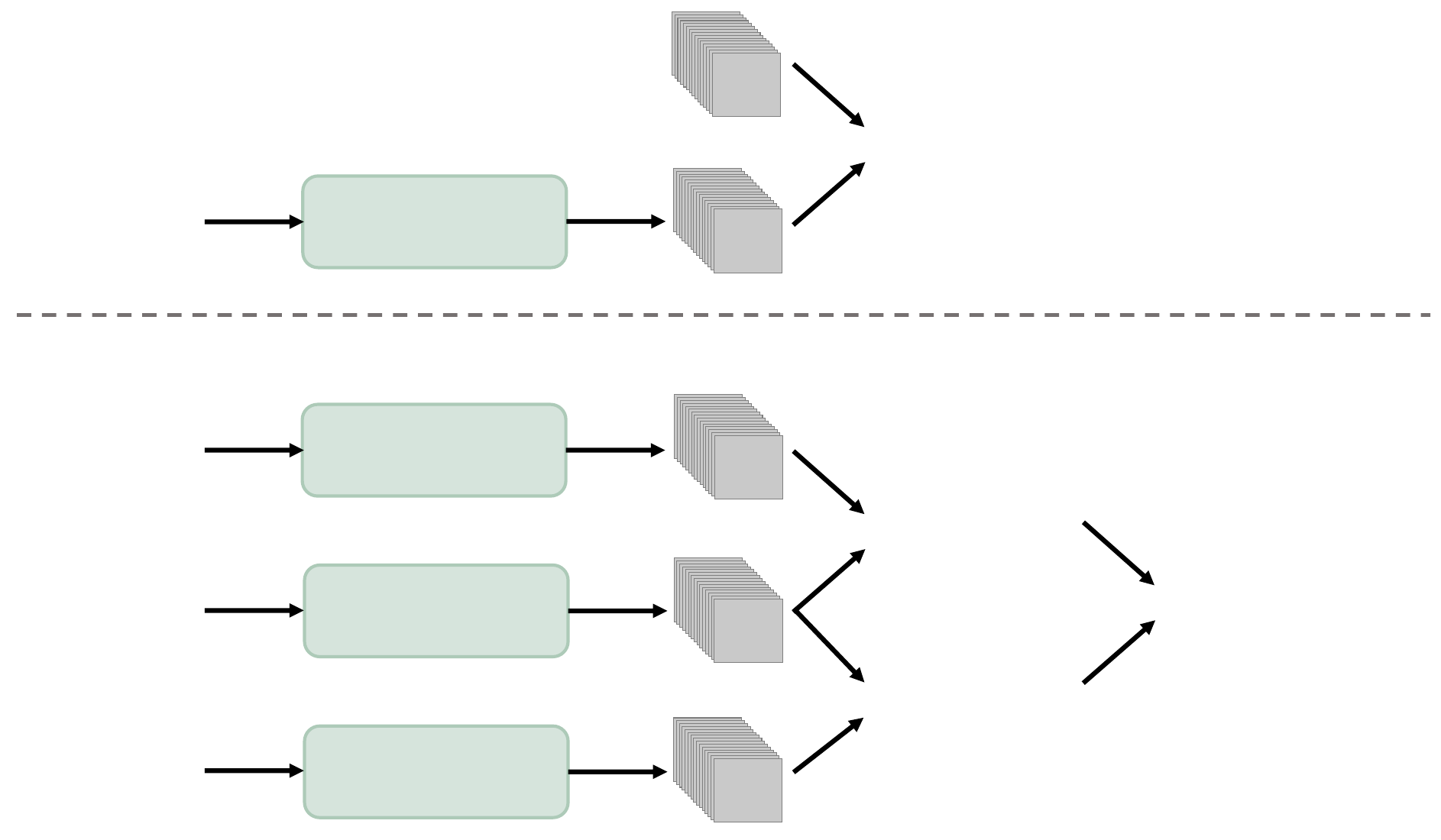}
{\small
\put(50.5,14){$p^{w}$}
\put(50.5,25.5){$p^{s_1}$}
\put(50.5,3){$p^{s_2}$}
\put(51.5,41.5){$p$}
\put(50,51.5){GT}
\put(62,9){$\HH(p^w, p^{s_2})$}
\put(62,21){$\HH(p^w, p^{s_1})$}
\put(82,16){$\mathcal{L}_{unlabeled}$} 
\put(29,4){$\FF$}
\put(29,15){$\FF$}
\put(29,26){$\FF$}
\put(29,42){$\FF$}
\put(7,15){$x^{w}$}
\put(7,26){$x^{s_1}$}
\put(7,4){$x^{s_2}$}
\put(7,43){$x$}
\put(62,48){$\mathcal{L}_{labeled} =  \mathcal{L}_{TP} + \mathcal{L}_{SoL} +\mathcal{L}_{Geo}$}
\put(0,57){Labeled Stream}
\put(0,32.5){Unlabeled Stream}
}
\end{overpic}
}
\caption{System overview of our proposed method. Here $\FF$ denotes the shared model, $x$ is labeled data, $x^w$ weakly perturbed unlabeled data, and $x^{s_{1,2}}$ strongly perturbed data. The loss function for the labeled data, $\mathcal{L}_{labeled}$ , 
is described in section \ref{sec:labeled}, and is directly built on the loss from \cite{mlsd}. For the unlabeled stream,  $\HH$ denotes the cross-entropy loss described in section \ref{sec:unlabeled} and is calculated between the weakly perturbed sample and one of the strongly perturbed samples.}
\label{fig:system}
\end{figure*}
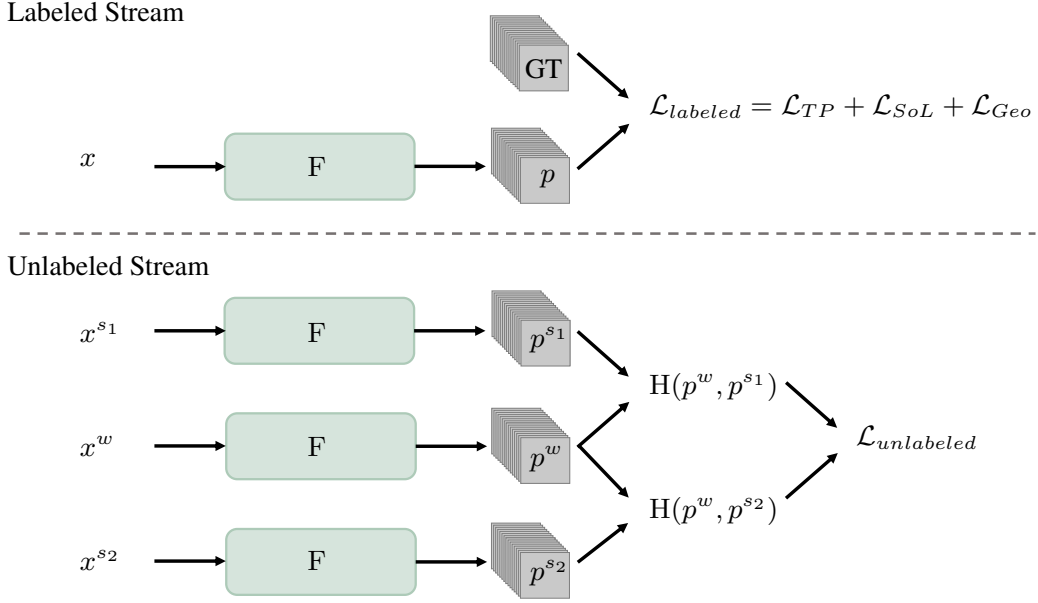
\section{A Semi-Supervised Line Segment Detector}
In this section, we describe the details of our semi-supervised system. The main idea is to show a deep learning network a small portion of annotated data to drive the network to detect lines, and at the same time show the network more data but without any annotations. To help the network draw information from the unannotated data, a stream with slightly different versions of the same image is presented and a consistency requirement on the detection added. We visualise the system in Fig. \ref{fig:system}, consisting of one part guided by existing ground truth, and one part utilizing information through unlabeled data. In our proposed method these two streams share the same model, $\FF$, which is guided by both the ground truth and the consistency requirement for the unlabeled data. 

The idea of using multiple strongly perturbed versions of the data is often used within classification \cite{caron2020unsupervised}, \cite{berthelot2019remixmatch}. It has also been   employed in semantic segmentation tasks\cite{unimatch}. There, the benefits of multiple strongly perturbed versions is explained as ''regularizing two strong views with a shared weak view can be regarded as enforcing consistency between these two strong views as well.'' \cite{unimatch}. We are therefore interested in transferring this idea into our unlabeled stream and investigate if this is favourable in line segment detection as well. We validate that it is, in our ablation studies.


In the following sections we will describe the model architecture and the different loss functions in detail.

\subsection{Model Architecture} \label{sec:model}

We want a fast, efficient and domain specific model that directly returns lines from an image. The framework M-LSD \cite{mlsd} meets these requirements. Utilizing the benefits of this fast line detector we build our semi-supervised framework for line segment detection with MobileNetV2 \cite{mobilenet} to gain an easily trained, fast model, using the same tiny model as in \cite{mlsd}. This model has around  0.6 million model parameters, and results in a model that runs in real time even on compute constrained devices \cite{mlsd}. 

The model is described by
\begin{equation}
    p = \FF(x),
\end{equation}
where  $x$ is the input RGB-image rescaled to size $512 \times 512 \times 3$ and $p$ is the output feature embedding of size $128 \times 128 \times 16$,  that codes the line representation. 

The network outputs 16 layers of feature maps. 
Seven of these are based on  the Tri-point (TP) line representation \cite{tp}. The TP is an over-parameterization of the line segment, coding it by the line segment center point and two displacement vectors (to the segment end points). The seven layers  consist of a length map, a degree map, four displacement maps, and one center map, all per pixel. The model uses the segment-of-line (SoL) augmentations, where each line is split into several TP representations. 
The next seven  feature maps are the TP-maps for the SoL and follow the same configuration as the  SoL feature maps. The last two feature layer maps are direct pixel-wise classification maps for line pixels and line junction pixels, respectively. All layers are used in the line loss, but only the first five feature layers are used at inference time for line detection.






\subsection{Labeled Loss} \label{sec:labeled}
 
The line segment loss for the labeled data
\begin{equation}
    \mathcal{L}_{labeled} = \mathcal{L}_{TP} + \mathcal{L}_{SoL} +\mathcal{L}_{Geo},
\end{equation}
 consists of three parts, a loss for the TP-representation $\mathcal{L}_{TP}$, a loss for the SoL augmentations, $\mathcal{L}_{SoL}$ and an additional geometric loss, $\mathcal{L}_{Geo}$ \cite{mlsd}. Here
\begin{equation}
    \mathcal{L}_{TP} = \mathcal{L}_{center} + \mathcal{L}_{disp} + \mathcal{L}_{match} ,
\end{equation}
where $\mathcal{L}_{center}$ is the weighted binary cross-entropy (WBCE) loss and $\mathcal{L}_{disp}$ is the L1 loss (as proposed in \cite{tp}). \cite{mlsd} formulates the $\mathcal{L}_{match}$ loss as
\begin{multline}
\mathcal{L}_{match  }= \frac{1}{|\mathbb{M}|} \sum_{(l, \hat{l}) \in \mathbb{M}}\left\|l_s-\hat{l}_s\right\|_1+\left\|l_e-\hat{l}_e\right\|_1 + \\+\left\|\bar{C}(\hat{l})-\left(l_s+l_e\right) / 2\right\|_1,
\end{multline}
where $\mathbb{M}$ is he set of matched line segments ($l, \hat{l}$) and $\bar{C}(\hat{l})$ is the center point of line $\hat{l}$ from the center map.
The $\mathcal{L}_{SoL}$ loss is set up identically to $\mathcal{L}_{TP}$, but for the segmented line representation.
The geometric information that \cite{mlsd} forms a loss for is four-fold. They formulate a segmentation loss as $\mathcal{L}_{seg} = \mathcal{L}_{junc} + \mathcal{L}_{line}$, where  both the loss terms are the WBCE loss. The last part of the loss functions is a regression loss, $\mathcal{L}_{reg} = \mathcal{L}_{length} + \mathcal{L}_{degree}$, where each loss term is a L1 loss. These two parts are then combined info one geometric loss, $\mathcal{L}_{Geo} = \mathcal{L}_{seg} + \mathcal{L}_{reg}$.

\subsection{Unlabeled Loss} \label{sec:unlabeled}
For the unlabeled data we adopt the method used in \cite{unimatch} for semantic segmentation, with some changes. For every unlabeled image three versions are loaded into the model, one weakly perturbed, and two different strongly perturbed versions. A consistency loss is then calculated between each of the two strongly perturbed versions of the image and the weakly perturbed version. The loss is formulated as
\begin{multline}
    \mathcal{L}_{unlabeled} = \frac{1}{N} \sum \mathbbm{1}(max(p^w) \geq\tau)(\text{H}(p^w, p^{s1}) +\\ + \text{H}(p^w, p^{s2})),
\end{multline}
where H is the cross-entropy loss and $\mathbbm{1}$ is the indicator function. The threshold $\tau$, makes sure that this loss only is applied when the prediction $p^w$ is certain enough. Since our model outputs 16 layers we chose to look at the layer representing certainty for the center point of the lines. 

The consistency loss is calculated over all of the output layers from the model, to ensure that the feature representation is kept throughout the training. 

\subsection{Augmentations} \label{sec:augmentations}

For the labeled data we use the same augmentations as in \cite{mlsd} consisting of random flip and rotation of the image, as well as the lines. A random hue, saturation, and value shift and a random brightness shift is also applied.

Following \cite{unimatch} for the augmentations the unlabeled data we first apply, a random flip and crop, which are recognized as the weak perturbations. Then, random blur, color jitter and gray scale are applied individually to two copies of the weakly perturbed image, the now strongly perturbed images.

Unlike the proposed method in \cite{unimatch} we do not have the feature augmented version of the unlabeled images, since we value a small and fast model, and chose to follow the model set up in \cite{mlsd} instead of having a larger bottle-neck model.

\paragraph{New CutMix version}
In \cite{unimatch} they make a strong motivation for using CutMix \cite{cutmix} for the unlabeled images and show that the models performance is boosted by this. One of the reasons is that this helps the model learn local image features better, without relying too much on full object or global image context.  However, for line segment detection we find that using the regular CutMix is unfavourable, and leads to a bias to very short line segments.
Therefore, we propose the following modification of the CutMix method. 
Instead of cutting out squares and patching them together for a more generalized model as in \cite{cutmix}, we only split the images along one dimension, either in x- or in y dimension with equal probability. This still helps with generalization but prevents the model from only seeing short lines.

\begin{table}
    \centering
     \caption{Comparison of the performance on the Finnwoodlands test dataset between our proposed method and the supervised one, for the different splits. The models here are trained only on the Finnwoodlands train dataset. The remaining data for each split were used as unlabeled data. The values reported are the mean values of three models, trained with identical settings. We also report results using the pre-trained models from \cite{xu2021line} and \cite{pautrat2023deeplsd}.}
    \label{tab:only_finn}
    \begin{tabular}{lrrrrr} \toprule
        Split  & ~ 1/16  & ~ 1/8~  & ~ 1/4~  & ~ 1/2~  & ~ 1 ~  \\
        \midrule
        {\scriptsize Supervised F$^H$} & 0.48 & 0.62 & 0.54 & 0.70 & 0.66 \\
         {\scriptsize Ours F$^H$} & \textbf{0.71} & \textbf{0.72} & \textbf{0.71} & \textbf{0.74} \\
         {\scriptsize LETR F$^H$} & & & & & 0.60 \\
       {\scriptsize DeepLSD F$^H$} & & & & & 0.64 \\
        \midrule
        {\scriptsize Supervised sAP$^{10}$} & 7.8 & 12.5 & 18.5 & 24.5 & 24.2 \\
         {\scriptsize Ours sAP$^{10}$} & \textbf{24.0} & \textbf{23.7} & \textbf{22.3} & \textbf{31.1}\\
       {\scriptsize  Gain ($\Delta$) } &  $\uparrow$ 16.2 & $\uparrow$ 11.2 & $\uparrow$ 3.8 & $\uparrow$ 6.6 \\
    {\scriptsize LETR sAP$^{10}$} & & & & & 9.2 \\
        \bottomrule 
    \end{tabular}
\end{table}

\section{Experiments} \label{sec:experiments}
In this section we will describe the experimental evaluation of the proposed framework.
In Sections~\ref{sec:datasets} and \ref{sec:settings} we 
describe the datasets and settings we used for the experiments. 
We carried out two kinds of experiment, \emph{Annotation dependence} and \emph{Domain generalization}.
In Section~\ref{sec:supervision_dependency} we conduct multiple test of how different portions of annotated data affect the resulting models. In Section~\ref{sec:domain_adaptation} we investigate how well we can use the proposed method to generalize a trained model to a new image domain without annotations. For both the tests in Section~\ref{sec:supervision_dependency} and \ref{sec:domain_adaptation} we compare the performances of our models to two pre-trained state-of-the-art (SOTA) models, DeepLSD \cite{pautrat2023deeplsd} and LETR \cite{xu2021line}. 
In Section \ref{sec:ablations} we conduct ablation studies for our proposed method. More images and extended ablation studies are provided in the supplementary material.
All training was done on an NVIDIA A100 Tensor Core GPU.

\subsection{Datasets and evaluation metrics} \label{sec:datasets}
We have used a number of datasets to test our method. The goal was to use these datasets to test two things. First, how well semi-supervised methods work in terms of the balance between labeled and unlabeled data. Second, how well image domain adaptation works, by using primarily unlabeled data for the new domain scenario. For these reasons we use one standard benchmark dataset of man-made scenes, \emph{the Wireframe dataset}, \cite{wireframe_cvpr18} and three domain specific datasets based on forestry imagery. 

The Wireframe dataset contains 5462 images, 5000 for training and 462 for test. Since we want a validation set during training we took out 300 images from the trainingset for this.

For the non man-made scenes we use \emph{the FinnWoodlands Dataset} \cite{finnwood} containing 300 RGB images, 250 for training and 50 for test, of snowy forest scenes in Finland, with corresponding segmentation masks for the trees, and surroundings. For the sake of our use we extract the silhouette lines of the trees from the segmentation masks. We split the validation set of 50 images into a validation set of 20 images and a testset of 30 images.

We also provide two new datasets of forestry scenes from two different forests, with annotated lines of the tree silhouettes, \skrylle\ and \snoge. \skrylle\ contains 975 images, where we used 775 for training and 200 images for test. \snoge\ contains 104 images and we used all these as an additional testset. We will make both these datasets publicly available. 

We evaluate our models performance using two metrics, heatmap-based F$^H$ \cite{end} and structural average precision with a threshold of 10 (sAP$^{10}$). For DeepLSD \cite{pautrat2023deeplsd} no direct confidence scores are available in the line detections and therefore we omit the sAP$^{10}$ score.

\subsection{Training Settings} \label{sec:settings}
For both the Finnwoodlands and Wireframe dataset, we create four different splits consisting of $1/2, 1/4, 1/8$ and $1/16$ part of the whole dataset with annotations. The rest of the dataset is used as unlabeled. 

As a first step we train a supervised model using only the labeled data, for each split, and the labeled loss. We ran the training for 300 epochs for Finnwoodlands and 200 epochs for Wireframe, respectively. The learning rate was set to 0.001, and through-out the training we keep the best model based on the sAP$^{10}$ score on the validation dataset. To form a baseline for comparison we made three identical runs for Finnwoodlands and two for Wireframe, reporting the mean performance. 

The weights from the respective supervised learning is then used for the semi-supervised training, lowering the learning rate to 0.0001 and training for another 100 epochs, again keeping the best model based on the sAP$^{10}$ score on the labeled validation dataset.

\subsection{Supervision dependency} \label{sec:supervision_dependency}
\begin{figure*}
  \centering
    \begin{tabular}{ccc}
  Fully Supervised~~~~~~~~~~~~~~~~~~~~~~~~~ & Proposed~~~~~~~~~~~~~~~~~~~~~~~~~~~~~~~~& Ground Truth~~~~~~~\\
  (Full dataset labeled)~~~~~~~~~~~~~~~~~~~~~~~~ &  (1/2 dataset labeled)~~~~~~~~~~~~~~~~~~~~~~~~~~~~~~~~~~~~ & 
  \end{tabular}\\
    \includegraphics[width=0.32\textwidth]{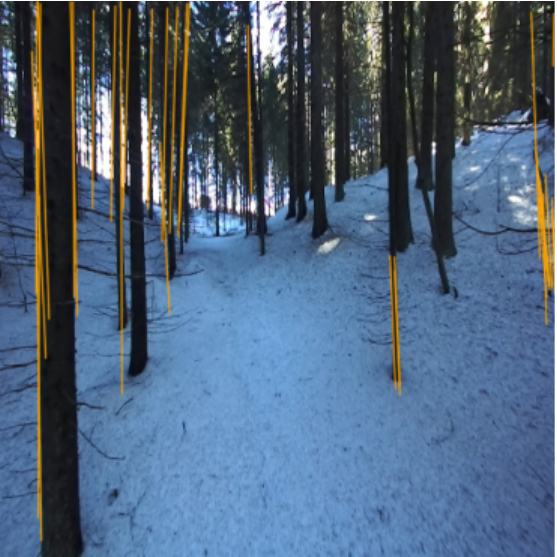}
    \includegraphics[width=0.32\textwidth]{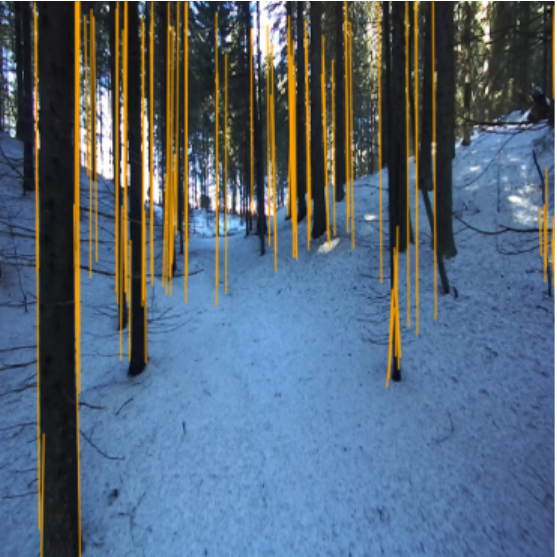}
    \includegraphics[width=0.32\textwidth]{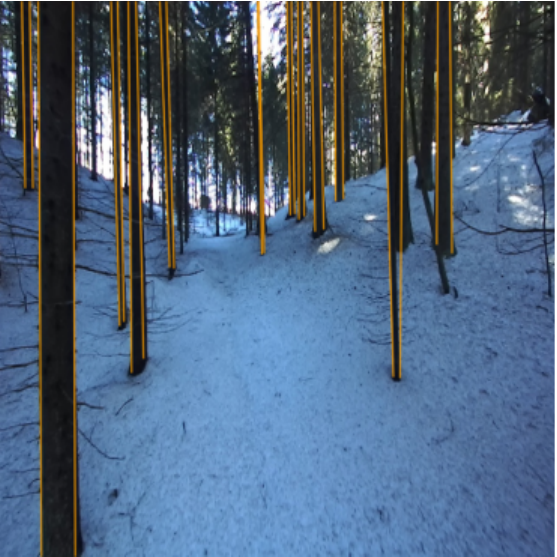}
  \caption{Comparison of the output results on one image from the Finnwoodlands test dataset. 
  From left to right, the fully supervised model, our proposed method trained with $1/2$ the dataset labeled, and  the ground truth. Both models are trained on the Finnwoodlands train dataset.}  
  \label{fig:finnwood}
\end{figure*}

\begin{figure*}
  \centering
   \begin{tabular}{ccc}
  Fully Supervised ~~~~~~~~~~~~~~~~~~~~~~~~ & Proposed~~~~~~~~~~~~~~~~~~~~~~ & ~~~~~~~~~~Ground Truth~~~~~~~~\\
  ~~~~(1/16 dataset labeled)~~~~~~~~~~~~~~~~~~~~~~~~~~~~ &  (1/16 dataset labeled)~~~~~~~~~~~~~~~~~~~~ & 
  \end{tabular}\\
    \includegraphics[width=0.32\textwidth]{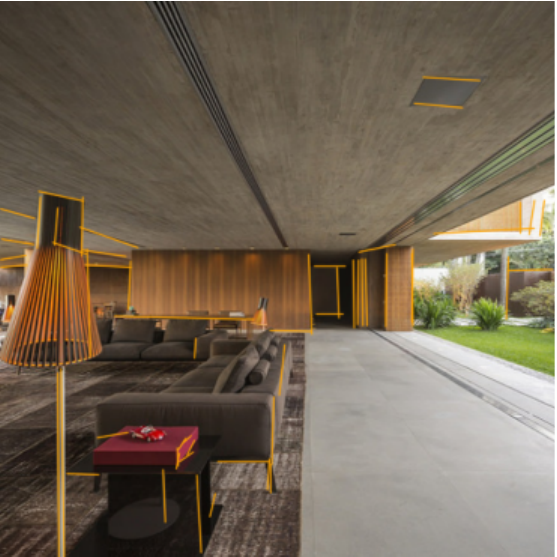}
    \includegraphics[width=0.32\textwidth]{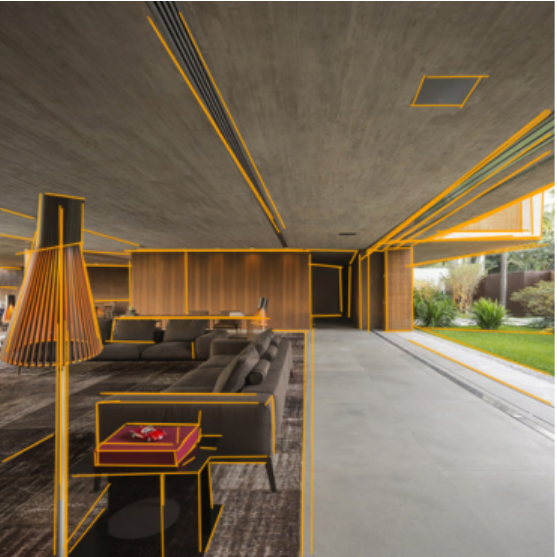}
    \includegraphics[width=0.32\textwidth]{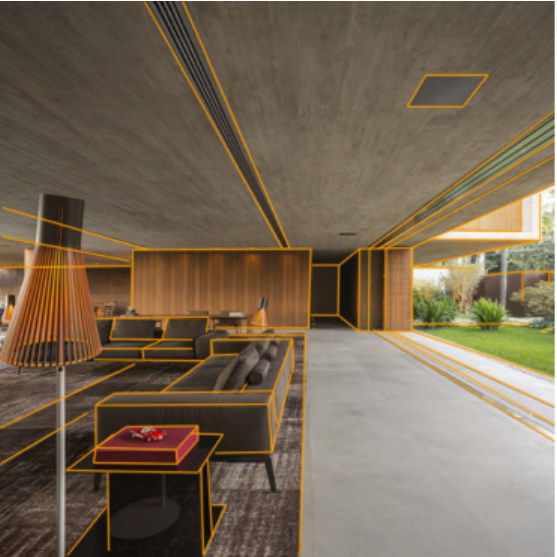}
  \caption{Comparison of the output results on one image from the Wireframe test dataset. 
  From left to right, the supervised model trained with $1/16$ the dataset labeled, our proposed method trained with $1/16$ the dataset labeled and the remaining as unlabeled, and  the ground truth.}
  \label{fig:wire}
\end{figure*}

In Table \ref{tab:only_finn} we demonstrate the performance of our proposed method in the case with not all of the data annotated. As shown, when adding unlabeled data and following our scheme, the results are better than only using the supervised model. Note also that for the $1/2$ split the performance exceeds the performance of the fully supervised model, visualized in Fig. \ref{fig:finnwood}. We see that our model finds more lines, without increasing the false positives excessively. Interestingly, for the $1/16$ split the performance of our model is similar to the fully supervised one, proving that the network is able to learn a great portion from the unlabeled stream. With sAP$^{10}$ as metric our method for the split $1/16$ is more than 200\% better than the supervised model for the same split. The performance of our models also exceeds the pre-trained SOTA models, demonstrating the possibility of creating an accurate model with very little data, and even less annotated data.
In addition to testing on the Finnwoodlands test dataset we also test the models performances on the \snoge\ test dataset, demonstrated in Table \ref{tab:snoge}, where we see that our method outperforms the fully supervised model on all the splits. This holds even for the model with only $1/16$ of the labeled data, as is visualized in Fig. \ref{fig:comparison1} where we note that our method allows for much greater generalization than the fully supervised. To emphasize this further, we compare the performance of our models to the SOTA models.
\begin{table*}
    \centering
    \caption{Comparison of the performance on the \snoge\ test dataset between our proposed method and the supervised one, for the different splits. The models here are trained only on the Finnwoodlands train dataset. The remaining data for each split were used as unlabeled data. The values reported are the mean values of three models, trained with identical settings.}
    \label{tab:snoge}
    \begin{tabular}{lccccc|cc} \toprule
        Split  & ~ 1/16 ~ & ~ 1/8 ~ & ~ 1/4 ~ & ~ 1/2 ~ & ~ 1 ~ & \multicolumn{2}{c}{Pretrained models} \\
       & & & & & &  LETR\cite{xu2021line} & DeepLSD\cite{pautrat2023deeplsd}  \\
        \midrule
        Supervised F$^H$ & 0.23 & 0.47 & 0.51 & \textbf{0.71} & 0.59 & 0.38 & 0.59 \\
        Ours F$^H$ & \textbf{0.68} & \textbf{0.52} & \textbf{0.71} & 0.69 &&&\\
        \midrule
        Supervised sAP$^{10}$ & 2.8 & 7.9 & 10.3 & 23.2 & 16.0 & 4.6 &\\
        Ours sAP$^{10}$ & \textbf{21.0} & \textbf{17.1} & \textbf{22.9} & \textbf{32.6} &&&\\
        Gain ($\Delta$)  &  $\uparrow$ 18.2 & $\uparrow$ 9.2 & $\uparrow$ 12.6 & $\uparrow$ 9.4 &&&\\
        \bottomrule 
    \end{tabular}
\end{table*}
In Table \ref{tab:wire} we demonstrate the performance of our method on a classic line segment detection dataset, the Wireframe dataset. We see that even on a larger dataset there are additional information that the network can learn with our method, compared to the supervised one. We visualize the detected lines in one image, for the $1/16$ labeled data case, in Fig. \ref{fig:wire}. Unsurprisingly LETR outperforms the other models, since this is the dataset LETR, and DeepLSD, are trained on. We got slightly lower sAP$^{10}$ score than previously reported for LETR, which is probably due to using other thresholds than they originally did. We have kept the threshold consistent through all of our experiments.

\begin{table*}
    \centering
    \caption{Comparison of the performance on the Wireframe test dataset between our proposed method and the supervised one, for the different splits. The remaining data were used as unlabeled data. The values reported are the mean values of two models, trained with identical settings}
    \label{tab:wire}
     \begin{tabular}{lccccc|cc}
    \toprule
        Split  & ~ 1/16 ~ & ~ 1/8 ~ & ~ 1/4 ~ & ~ 1/2 ~ & ~ 1 ~ & \multicolumn{2}{c}{Pretrained models} \\
       & & & & & &  LETR\cite{xu2021line} & DeepLSD\cite{pautrat2023deeplsd}  \\
        \midrule
        Supervised F$^H$ & 0.75 & 0.76 & \textbf{0.76} & 0.76 & 0.76 & 0.83 & 0.72 \\
        Ours F$^H$ & 0.75 & 0.76 & 0.75 & 0.76 & & & \\
        \midrule
        Supervised sAP$^{10}$  & 23.1 & 32.9 & 41.8 & 49.3 & 53.7 & 50.5 &\\
        Ours sAP$^{10}$ & \textbf{34.0} & \textbf{40.0} & \textbf{44.9} & \textbf{50.1} & \\
        Gain ($\Delta$) & $\uparrow$ 10.9 & $\uparrow$ 7.1 & $\uparrow$ 3.1 & $\uparrow$ 0.8 &\\
        \bottomrule 
    \end{tabular}
\end{table*}

\begin{table*}
    \centering
    \caption{Comparison of the performance on the \skrylle\ test dataset between the supervised model, our proposed method with only the remaining data from Finnwoodlands training dataset as unlabeled data, and our method with the \skrylle\ train dataset without labels, for the different splits. The models here are trained only on labeled data from the Finnwoodlands train dataset.}
    \label{tab:finn_skrylle}
    \begin{tabular}{lccccc|cc}
    \toprule
        Split  & ~ 1/16 ~ & ~ 1/8 ~ & ~ 1/4 ~ & ~ 1/2 ~ & ~ 1 ~ & LETR & DeepLSD\\
        \midrule
        Supervised F$^H$ & 0.16 & 0.40 & 0.39 & 0.38 & 0.35 & 0.23 & 0.38\\
        Ours F$^H$ & 0.35 & 0.38 & 0.31 & \textbf{0.46} & && \\
        Ours with \skrylle\ F$^H$ &\textbf{ 0.40} & \textbf{0.42} & \textbf{0.44} & 0.33 & \textbf{0.40} && \\
        \midrule
        Supervised sAP$^{10}$  & 2.1 & 4.2 & 5.5 & 4.4 & 6.0 & 2.3 &\\
        Ours sAP$^{10}$ & \textbf{9.8} & 10.0 & 3.9 & \textbf{14.4} &&&  \\
        Ours with \skrylle\ sAP$^{10}$ & 6.5 & \textbf{15.4} & \textbf{14.2} & 11.4 & \textbf{13.8}&&\\
        \bottomrule 
    \end{tabular}
\end{table*}

\begin{table}[H]
    \centering
    \caption{Ablation study for the Finnwoodlands dataset with 1/2 the dataset labeled, tested on Finnwoodlands testset. Again each test is trained three times, with identical setting. The reported performance is the mean of the three.}
    \label{tab:ablation_finn}
 \begin{tabular}{lcc}
    \toprule
    &  F$^H$ &  sAP$^{10}$ \\
     \midrule
    A single $x^s$ & 0.72 $\pm 0.029$ & 28.2 $\pm 0.81$ \\
    CutMix\cite{unimatch} & 0.72 $\pm 0.040$  &   33.8 $\pm 5.89$ \\
    w/o CutMix &  0.72 $\pm 0.046$ &     29.5 $\pm 3.44$ \\
    Proposed  &  0.74 $\pm 0.015$ &   31.1 $\pm 2.72$\\
 \bottomrule 
    \end{tabular}
\end{table}

\subsection{Domain Adaptation} \label{sec:domain_adaptation}
In addition to the first experiments we also look at how the performance of a model on a specific, previous unseen, dataset can be boosted by adding data from that dataset without any annotation, investigating the frameworks ability to improve the model for a domain specific case. For this experiment we used the splits of Finnwoodlands with annotations and added another dataset, the completely unlabeled trainingset of \skrylle. We evaluate the results on the testset of \skrylle. 

We show the results of this experiment in Table \ref{tab:finn_skrylle}. Here we provide both the results of our method trained on only Finnwoodland data, and our method that is trained with the labeled data from Finnwoodlands and the \skrylle\ train dataset as unlabeled data. In most of the splits our method with the \skrylle\ train dataset as unlabeled data performs the best, and in the cases it doesn't our regular model outperforms the supervised one. As demonstrated by both previous tests and this, the pretrained SOTA models fail to perform well on a novel domain.



\subsection{Ablation Studies} \label{sec:ablations}

To analyze our proposed method we conduct several ablation studies. Here we report the results for $1/2$ labeled data. 
We demonstrate in Table \ref{tab:ablation_finn} the benefits of having two strongly perturbed versions of the unlabeled images instead of one. We also illustrate the large variance that comes from using the original CutMix \cite{cutmix}, and the benefit of having our CutMix compared to not having any CutMix. All of these models are trained and tested on the Finnwoodlands dataset.

\begin{figure*}[ht!]
  \centering
  \begin{tabular}{ccc}
  Proposed CutMix ~~~~~~~~~~~~~~~~~~~~& Original CutMix & ~~~~~~~~~~~~~~~~~~~~Ground Truth~~~~~~~~\\
  ~~~~(1/2 dataset labeled)~~~~~~~~~~~~~~~~~~~~~~~~ &  ~~~~~~~~(1/2 dataset labeled)~~~~~~~~ & 
  \end{tabular}\\
    \includegraphics[width=0.32\textwidth]{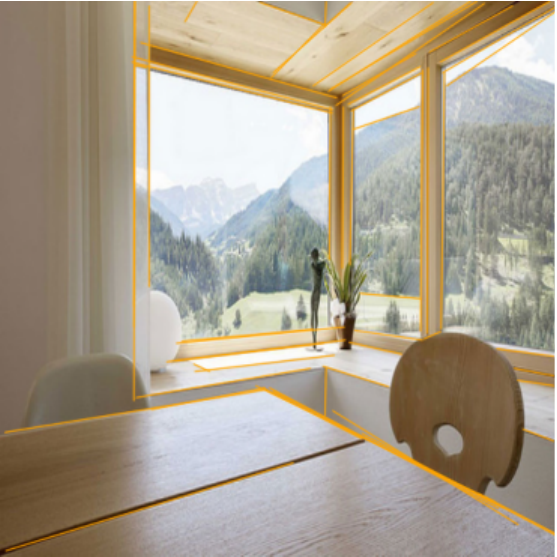}
    \includegraphics[width=0.32\textwidth]{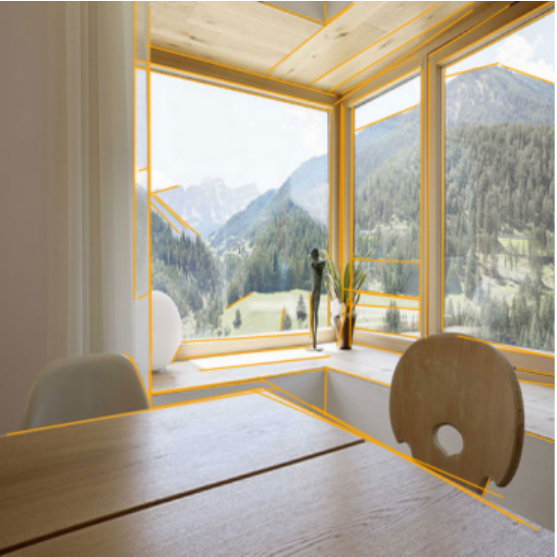}
    \includegraphics[width=0.32\textwidth]{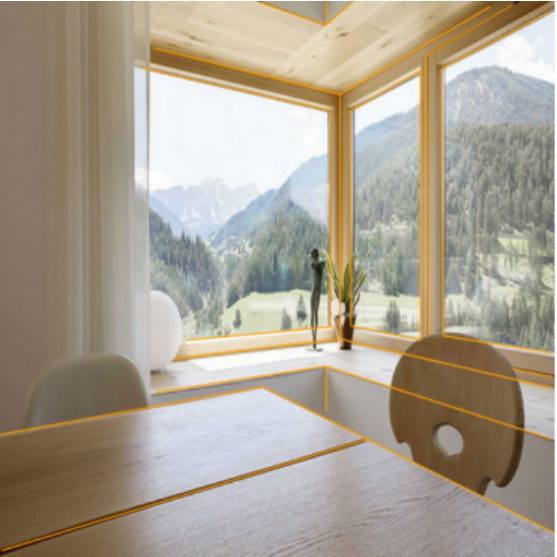}
    \includegraphics[width=0.32\textwidth]{figs/unimatch_1_2_finn_20_croped.pdf}
    \includegraphics[width=0.32\textwidth]{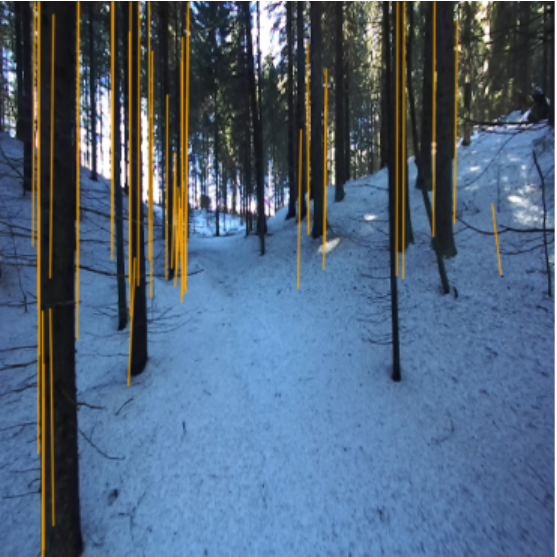}
    \includegraphics[width=0.32\textwidth]{figs/ground_truth_finn_20_croped.pdf}
  \caption{Comparison between our proposed CutMix and the original CutMix \cite{cutmix}, using our training scheme with  $1/2$ the dataset as labeled and the remaining as unlabeled. 
  Top row shows an example from the Wireframe dataset, and bottom 
  from the Finnwoodlands dataset. Notice how the original CutMix has more short lines.}
  \label{fig:cutmix}
\end{figure*}

To further demonstrate the benefit of our CutMix compared to the original one, and to show that ours is more suitable for lines in all kinds of scenes we make the same experiment on the Wireframe dataset, shown in Table \ref{tab:ablation_wire}. In Fig \ref{fig:cutmix} we visualize some outputs from our proposed CutMix alongside the original CutMix. Here the original CutMix detects  more short and inaccurate lines than the proposed version.


\begin{table}[H]
    \centering
     \caption{Ablations study for the Wireframe dataset with 1/2 the dataset labeled, tested on Wireframe testset. Again each test is trained two times, with identical setting. The reported performance is the mean of the two.}
    \label{tab:ablation_wire}
     \begin{tabular}{lcc}
    \toprule
          & org. CutMix\cite{unimatch} & Proposed \\
        \midrule
        F$^H$  & 0.75 $\pm 0.014$&  0.76 $\pm 0.007$\\
        sAP$^{10}$ & 50.3 $\pm 0.57$ &   50.1 $\pm 0.50$\\
        \bottomrule 
    \end{tabular}
\end{table}

To ensure that the semi-supervised scheme is not simply a favourable training setting, we let the model train on the whole Finnwoodlands dataset, both considered as labeled and unlabeled data. Thus letting the network see all the images, and getting feedback from the ground truth, while at the same time utilizing the augmentations of the unlabeled training stream, making sure that the consistency requirement is met. In Table \ref{tab:unimatch_all} we see that this does, however, not result in a better performing model than the one obtained with the $1/2$ split, which only sees half the dataset with ground truth guidance and the other half with the consistency requirement on. 

\begin{table}[H]
    \centering
    \caption{Comparison between using half the training dataset with labels and half without labels, and using the whole dataset with labels, and the same whole dataset as unlabeled. The reported values are the mean of three identical runs.}
    \label{tab:unimatch_all}
     \begin{tabular}{lcc}
    \toprule
          & ~ Finnwoodlands ~ & ~ \snoge ~ \\
        \midrule
        F$^H$ Split ($2/1$) & 0.73 & \textbf{0.71 }\\
        F$^H$ Split ($1/2$) & \textbf{0.74} & 0.69 \\
        \midrule
        sAP$^{10}$ Split ($2/1$)& 30.2 & 29.8 \\
        sAP$^{10}$ Split ($1/2$)& \textbf{31.1} & \textbf{32.6} \\
        \bottomrule
    \end{tabular}
\end{table}

\section{Conclusion}
We have in this paper presented the first framework for semi-supervised learning for line segment detection. Through extensive testing on diverse datasets, we demonstrate that these methods improve significantly on the state-of-the-art for new environments or  where there is little annotation. Furthermore, we show that small models, suitable for real-time applications,  benefit greatly from semi-supervised methods, and indeed in some cases outperforming methods supervised on the full dataset. Future work includes testing these methods on various down stream tasks as well as adapting the model to handle other low parametric features than lines.
{
    \small
    \bibliographystyle{ieeenat_fullname}
    \bibliography{main}
}


\end{document}